\documentclass[letterpaper, 10 pt, conference]{ieeeconf}  

\IEEEoverridecommandlockouts                              

\usepackage{graphics} 
\usepackage{graphicx}
\usepackage{booktabs}
\usepackage{multirow}
\usepackage{cite}
\usepackage{amsmath,amssymb,amsfonts}
\usepackage{textcomp}
\usepackage{xcolor}
\usepackage{caption}
\usepackage{subcaption}
\usepackage[ruled,vlined]{algorithm2e}
\usepackage{lipsum}
\usepackage{gensymb}

\usepackage{amsthm}
\usepackage{float}
\let\labelindent\relax
\usepackage[inline]{enumitem}
\usepackage[breaklinks,colorlinks]{hyperref}
\usepackage[english]{babel}

\usepackage[utf8]{inputenc}
\usepackage{inconsolata} 
\usepackage[most]{tcolorbox}
\tcbuselibrary{skins,breakable}

\newtcblisting{promptcode}[2][]{%
  enhanced,
  listing engine=listings,
  listing only,
  colback=gray!7, colframe=black, boxrule=0.5pt, arc=1.2mm,
  left=1mm, right=1mm, top=1mm, bottom=1mm,
  fonttitle=\bfseries, title={#2},
  listing options={
    basicstyle=\ttfamily\small,
    breaklines=true, breakatwhitespace=true,
    breakautoindent=false, breakindent=0pt,
    columns=fullflexible, keepspaces=true
  },
  before upper=\ignorespaces,
  after  upper=\unskip\ignorespacesafterend,
  #1
}

\newcommand{\obs}{\ensuremath{\mathbf{o}}}
\newcommand{\objset}{\ensuremath{\mathcal{O}}}
\newcommand{\act}{\ensuremath{\mathbf{a}}}
\newcommand{\lan}{\ensuremath{\mathbf{l}}}
\newcommand{\pick}{\ensuremath{\mathcal{T}_{\text{pick}}}}
\newcommand{\place}{\ensuremath{\mathcal{T}_{\text{place}}}}
\newcommand{\atol}{\ensuremath{a \rightarrow l}}
\newcommand{\ltoa}{\ensuremath{l \rightarrow a}}

\newcommand{\ltoc}{\ensuremath{l \rightarrow c}}

\newcommand{\demoset}{\ensuremath{\mathcal{D}}}
\newcommand{\demo}{\ensuremath{\zeta}}
\newcommand{\policy}{\ensuremath{\pi}}
\newcommand{\LtoA}{\ensuremath{\text{L2A}}}
\newcommand{\AtoL}{\ensuremath{\text{A2L}}}

\newcommand{\LtoC}{\ensuremath{\text{L2C}}}
\newcommand{\LtoAtoL}{\ensuremath{\text{L2A2L}}}

\title{\LARGE \bf
LACY: A Vision-Language Model-based Language-Action Cycle for Self-Improving Robotic Manipulation}

\author{Youngjin Hong$^*$, Houjian Yu$^*$, Mingen Li, and Changhyun Choi\\
\thanks{$^*$Equal contribution. Author order is alphabetical. This work was supported in part by the Sony Research Award Program and NSF Award 2143730. 
The authors are with the Department of Electrical and Computer Engineering, Univ. of Minnesota, Minneapolis, USA
        {\tt\small \{hong0745, yu000487, li002852, cchoi\}@umn.edu}}}

\begin{document}\maketitle
\thispagestyle{empty}
\pagestyle{empty}

\begin{abstract}
Learning generalizable policies for robotic manipulation increasingly relies on large-scale models that excel at mapping language instructions to actions ($\LtoA$). However, this unidirectional training paradigm often produces policies that can execute tasks without deeper contextual understanding, thereby limiting their ability to generalize and to explain their behavior. We argue that the complementary skill of mapping actions back to language ($\AtoL$) is essential for developing more holistic and robust grounding. An agent capable of both acting and explaining its actions can form richer internal representations and, critically, unlock new paradigms for self-supervised learning.
In this paper, we introduce LACY (Language-Action CYcle), a unified framework that learns such bidirectional mappings within a single vision-language model. LACY is jointly trained on three synergistic tasks: generating parameterized actions from language ($\LtoA$), explaining observed actions in language ($\AtoL$), and verifying semantic consistency between two language descriptions ($\LtoC$). The framework enables a self-improving cycle that autonomously generates new training data by chaining the $\LtoA$ and $\AtoL$ modules in an $\LtoAtoL$ pipeline. The $\LtoC$ module then filters this data using an active data augmentation strategy that selectively targets low-confidence cases, thereby improving the model efficiently without requiring additional human annotations. Extensive experiments on pick-and-place tasks in both simulation and the real world demonstrate that LACY substantially improves task success rates by 50.56 percentage points on average compared to baseline methods and yields more robust language-action grounding for robotic manipulation. For more details, please refer to our project page:~\url{https://vla2026.github.io/LACY/}
\end{abstract}

\section{Introduction}

\begin{figure}[t]
    \centering
    \includegraphics[width=\columnwidth]{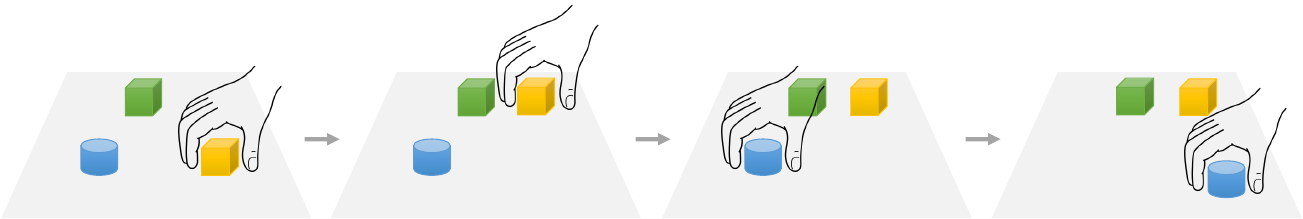}
    \vspace{-12pt}
    \caption{\textbf{Human demonstration of toy object manipulation.} Humans can readily infer task procedures from a manipulation demonstration and express them in language (e.g., ``pick up the yellow block'' $\rightarrow$ ``place it to the right of the green block'' $\rightarrow$ ``grasp the blue cylinder'' $\rightarrow$ ``put it on the bottom right of the table''). This linguistic description enables humans to accurately replicate the demonstrated action sequence.}
    \label{fig:intro}
\end{figure}

Human actions and language are deeply and bidirectionally intertwined~\cite{glenberg_grounding_2002,rizzolatti_language_1998,pulvermuller_brain_2005}. 
Humans can rapidly acquire new tasks by observing demonstrations, semantically interpreting the underlying task procedures, and describing the observed actions in natural language. For example, after watching the toy block manipulation shown in Fig.~\ref{fig:intro}, one can readily articulate the task as a sequence of sentences. Conversely, when given a textual task description, humans can typically (1) interpret the intended actions (e.g., pick, place, grasp, put), (2) ground the mentioned objects in the environment (e.g., yellow block, blue cylinder), and (3) execute the described action sequence accordingly.

Learning from Demonstration (LfD) and imitation learning have been extensively studied for robotic manipulation tasks~\cite{jang2022bc, zhao2023learning, chi2023diffusion, ze20243d, shridhar2023perceiver}. While these approaches have demonstrated strong performance, they typically require large amounts of high-quality demonstration data, which are costly and labor-intensive to collect~\cite{zeng2021transporter}. Recently, the emergence of Vision-Language Models (VLMs) has driven substantial progress by leveraging their generalization capabilities and open-world reasoning. However, existing approaches, from early framework such as CLIPort~\cite{shridhar2022cliport} to large-scale generalist models like RT-2~\cite{brohan2023rt2} and OpenVLA~\cite{kim2024openvla}, have predominantly focused on a single paradigm: learning a unidirectional mapping from \textit{language-to-action} ($\LtoA$). Although these $\LtoA$ approaches have achieved notable success, their effectiveness remains fundamentally limited by their dependence on massive, passively collected datasets, which constrains both scalability and data efficiency.

To address this fundamental limitation, we advocate for an approach grounded in a more holistic, bidirectional integration of language and action. Inspired by findings in neuroscience~\cite{rizzolatti_language_1998,pulvermuller_brain_2005}, we posit that the complementary ability to describe tasks by observing manipulation actions, referred to as \textit{action-to-language} ($\AtoL$), is equally essential yet remains relatively underexplored in robot learning. An agent equipped with both $\LtoA$ and $\AtoL$ capabilities can develop richer and more robust internal representations of the world. More critically, this synergy enables a powerful self-supervised learning paradigm: the model can execute a robot action from a language command ($\LtoA$), then generate a new language description of that action based on its own perception ($\AtoL$), and finally assess whether the initial and generated descriptions are semantically consistent ($\LtoC$). This cycle allows the model to generate, evaluate, and refine its own data, thereby substantially reducing its reliance on external human supervision.

In this paper, we introduce \textbf{LACY} (Language-Action CYcle), a unified VLM framework that leverages the language-action cycle for robust and scalable learning. LACY is implemented by fine-tuning a single LLaVA-NeXT model~\cite{liu2024llavanext} to perform three complementary yet synergistic roles: (1) generating actions from language ($\LtoA$), (2) explaining actions in language ($\AtoL$), and (3) verifying semantic consistency between two language descriptions ($\LtoC$). We first train the $\LtoA$ and $\AtoL$ components on an initial dataset and then use the proposed language–action cycle to autonomously generate a large-scale, high-quality dataset for further training. A central innovation of our framework is a confidence-based active data augmentation strategy, in which new data is stochastically generated only for low-confidence scenarios, thereby directing learning toward more challenging cases. We evaluate LACY on a tabletop pick-and-place task in both simulation and real-world settings. Experimental results show that policies trained through LACY’s self-improvement loop significantly outperform baselines trained solely on the initial dataset, yielding substantial gains in task success rates.

The main contributions of this work are as follows:
\begin{itemize}
    \item \textbf{A unified VLM framework (LACY)} that is jointly trained to perform three complementary tasks: language-to-action generation ($\LtoA$), action-to-language explanation ($\AtoL$), and semantic consistency verification ($\LtoC$).
    \item \textbf{A self-improving data generation pipeline} that leverages the language-action cycle to autonomously produce new training data, which is subsequently filtered by the $\LtoC$ module to ensure high-quality data.
    \item \textbf{A confidence-based active data augmentation strategy} that directs data generation toward low-confidence scenarios, thereby mitigating overfitting and improving model performance on challenging cases.
\end{itemize}

\section{Related Work}

\subsection{Vision-Language Models for Robotic Manipulation}
One of the challenges for generalist robot policies lies in grounding vision, language, and action, which exhibit significant abstraction gaps. Outside of robotics, there have been many successful attempts to bridge the vision and language feature spaces~\cite{li2023blip, zhai2023sigmoid}. Along with the success of vision-based imitation learning~\cite{jang2022bc, zhao2023learning}, early attempts were made in language-grounded imitation learning~\cite{yang2022interactive, yu2025parameter}. 

With the advent of state-of-the-art VLM models~\cite{liu2023visual, bai2023qwen}, recent works tend to directly leverage the generalizability and reasoning capabilities of VLMs for robotic tasks. One dominant paradigm is end-to-end Vision-Language-Action (VLA) models~\cite{shridhar2022cliport, kim2024openvla}. Given language instructions and image observations, VLA models directly generate robot actions. Thanks to large-scale robot datasets~\cite{walke2023bridgedata, liu2023libero}, VLA models are built upon existing open-source VLM models by fine-tuning with robot datasets. Recent works also explore diverse approaches for faster and high-performance fine-tuning and inference of VLA models~\cite{pari2024finetuning, pertsch2025fast}. However, due to the lack of robot data compared to open-world data from which VLMs were trained, these models must be trained on additional in-domain robot data, which are expensive and require significant time, equipment, and expertise to collect.

Recent approaches explore hierarchical VLA architectures that use VLMs to generate intermediate action representations and employ existing low-level policies~\cite{yuan2024robopoint, nasiriany2024}. By using abstract robot action expressions, hierarchical models can fully leverage the reasoning capabilities of VLM models, either by fine-tuning with off-domain data~\cite{li2025hamster}, or even through zero-shot inference via off-the-shelf VLMs~\cite{qian2024thinkgrasp}. Our work has connections with these prior methods, as our L2A task generates a parameterized pick-and-place motion.

Prior works have mainly focused on how to collect or utilize large-scale data for robotic tasks. However, these models are almost exclusively trained for the unidirectional $\LtoA$ task, while the complementary skill of observing an action and generating a linguistic description, $\AtoL$, remains largely unexplored. We argue that learning the A2L approach deepens understanding of robot tasks, and this can maximize learning efficiency from limited in-domain demonstrations through bidirectional grounding.

\subsection{Data Generation and Self-Supervision in Robotics}
The immense data requirements of VLA models have spurred research into data generation and self-supervision. Common strategies include using simulation to create large-scale datasets~\cite{mandlekar2023mimicgen, liu2023libero}, though this often introduces sim-to-real gaps. Reinforcement learning approaches have also been explored to enable autonomous data collection and policy improvement~\cite{kumar2021rma, kalashnikov2018scalable}, but these methods typically require extensive environment interaction and careful reward design. Alternative approaches focus on enabling models to learn from their own experience through self-supervision~\cite{burda2018exploration, srinivas2020curl}, data augmentation technique~\cite{kostrikov2020image}, and cross-modal transfer~\cite{sermanet2018time}. However, a key challenge in self-supervised learning is ensuring the quality of self-generated data, as models may reinforce their own biases without external verification~\cite{huang2023large, valmeekam2023can}.

LACY provides a novel form of self-supervision by framing the problem through the lens of cycle consistency, a powerful principle from generative modeling~\cite{zhu2017unpaired}. The core idea is that a forward mapping followed by a backward mapping should reconstruct the original input. This has been applied in robotics for tasks like cross-domain policy transfer~\cite{pmlr-v164-ajay22a}, but its application to the language-action domain for data generation is new. Our language-action cycle ($\lan \rightarrow \act \rightarrow \lan'$) provides a principled self-supervisory signal as the semantic intent of the initial command $\lan$ must be preserved in the reconstructed description $\lan'$. The $\LtoC$ module acts as the crucial verifier, providing the reliable, extrinsic-like feedback that naive self-correction lacks. By combining this with a confidence-based active data augmentation strategy, inspired by methods for calibrating LLM uncertainty~\cite{kadavath2022language}, LACY focuses its learning on ambiguous cases, making the self-improvement process both robust and efficient.

\section{Proposed Approach}

\begin{figure}[t]
  \begin{flushright}
    \includegraphics[width=0.48\textwidth]{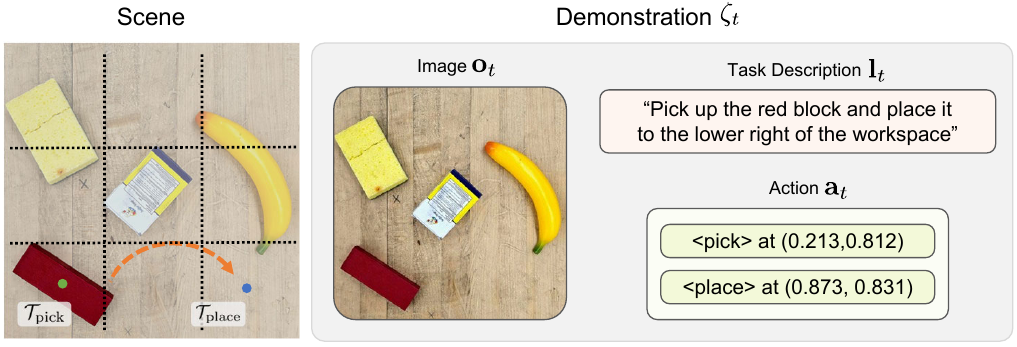}
    \vspace{-12pt}
    \caption{\textbf{Notations.} Each demonstration $\demo_t$ includes an image observation $\obs_t$, a task description $\lan_t$, and a pick-and-place action $\act_t$. The workspace is divided into a $3\times3$ grid. Coordinates $(x, y)$ are normalized to $[0,1]$, where $x,y\in[0,1]$, with $(x,y)=(0,0)$ at the left/top image border and $(x,y)=(1,1)$ at the right/bottom border.}
    \label{fig:notation}
  \end{flushright}
\end{figure}

\subsection{Preliminary: Problem Definition and Notations}
Following the notations in~\cite{zeng2021transporter, shridhar2022cliport}, we assume that a set of $n$ demonstrations $\demoset = \{ \demo_1, \demo_2, \cdots, \demo_n \}$ are given. Each demonstration $\demo_t = (\obs_t, \lan_t, \act_t)$ is a triplet containing an image observation $\obs_t$, a task description in human language $\lan_t$, and a pick-and-place action $\act_t = (\pick, \place) \in \mathbb{R}^{2}\times\mathbb{R}^{2}$, as shown in Fig.~\ref{fig:notation}.

From the demonstration set $\demoset$, a language-to-action ($\LtoA$) model learns a policy $\policy_{\ltoa}$ that maps an observation $\obs_t$ and a language instruction $\lan_t$ to a predicted action $\hat{\act}_t$:
\begin{equation} 
    \policy_{\ltoa}: \obs_t, \lan_t \rightarrow \hat{\act}_t.
\end{equation} 
Conversely, our proposed action-to-language ($\AtoL$) model learns a policy $\policy_{\atol}$ that predicts a language description $\hat{\lan}_t$ from an observation $\obs_t$ and a demonstrated action $\act_t$:
\begin{equation} 
    \policy_{\atol}: \obs_t, \act_t \rightarrow \hat{\lan}_t.
\end{equation} 

The language $\lan_t$ of a pick-and-place action $\act_t$ includes which object to pick and where to place, for which absolute or relative spatial description is employed (see Section~\ref{sec:approach:atol} for details).

Additionally, we introduce a language-to-consistency ($\LtoC$) model that estimates the consistency $c_t \in [0, 1]$ between two languages ($\lan_t$ and $\hat{\lan}_{t}$) conditioned on the observation $\obs_t$:
\begin{equation} 
    \policy_{\ltoc}: \obs_t, \lan_t, \hat{\lan}_t \rightarrow c_t.
\end{equation} 

\begin{figure}[tbh]
    \begin{center}
        \includegraphics[width=0.6\columnwidth]{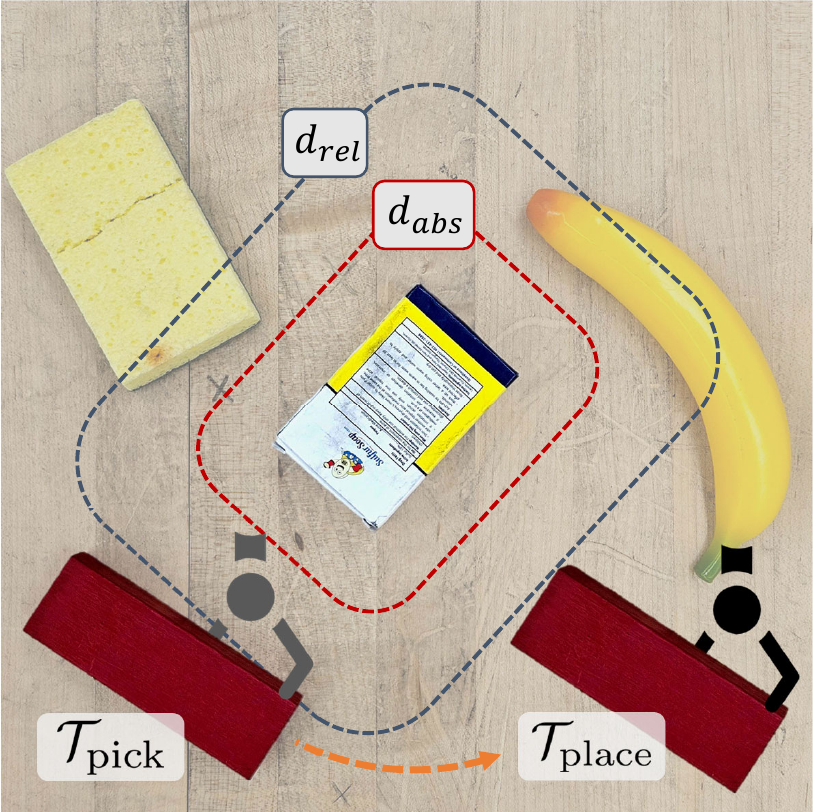}%
    \end{center}
    \vspace{-10pt}
    \caption{\textbf{Spatial description types.} Task description for placing an object uses different forms of language descriptions (absolute or relative) based on the Euclidean distance to the placing location and the proximity to the outer contour of the nearest object.}
    \label{fig:exman_thresholding}
\end{figure}

\begin{figure*}
    \includegraphics[width=1\textwidth]{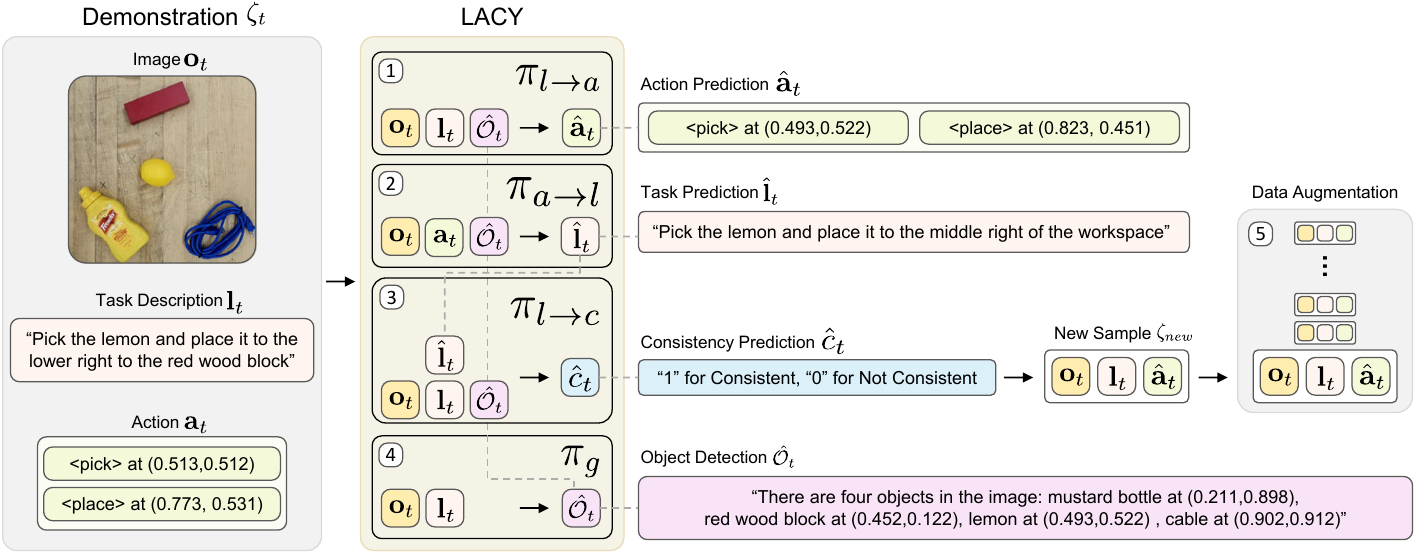}
    \vspace{-14pt}
    \caption{\textbf{Overview of the LACY framework.} LACY (Language-Action CYcle) builds upon a single VLM~\cite{liu2024llavanext} fine-tuned to serve three roles: (1) an action generator ($\LtoA$), (2) an action explainer ($\AtoL$), and (3) a consistency verifier ($\LtoC$). The framework operates as a closed-loop system, where these bidirectional capabilities enable LACY to generate new high-quality training data and iteratively refine itself. (4) Each task is framed as a chain-of-thought (CoT) process, where the model first performs object grounding to predict object names and locations ($\hat{\objset}$) and then uses this contextual information to complete the target task. (5) As shown in Algorithm \ref{alg:checkman}, new samples are generated and merged with the initial dataset.}
    \label{fig:overview}
\end{figure*}

\subsection{System Overview}
We introduce LACY (Language-Action CYcle), a framework built upon a single, powerful VLM (LLaVA-NeXT~\cite{liu2024llavanext}) that is fine-tuned to serve three roles: an action generator ($\LtoA$), an action explainer ($\AtoL$), and a consistency verifier ($\LtoC$). Fig.~\ref{fig:overview} illustrates the overall architecture. The core of our framework is a closed-loop process where LACY uses its bidirectional capabilities to generate new, high-quality training data, which is then used to iteratively refine the model itself.

\subsection{Explaining Robot Actions in Language via $\AtoL$}
\label{sec:approach:atol}
The main goal of the $\AtoL$ model is to generate a language description by observing a manipulation action. Since we consider pick-and-place actions, the language needs to describe which object to grasp and where to place it, which necessitates spatial description. We designed $\AtoL$ to generate two distinct types of naturalistic spatial descriptions: absolute and relative. Instead of using unconstrained free-form descriptions, we constructed a controlled set of linguistic templates to describe robot actions. This provides consistency across samples while maintaining sufficient linguistic diversity for generalization. From an image $\obs_t$ and an action $\act_t$, we generate language based on the spatial context.

\begin{itemize}
    \item \textbf{\textit{Absolute Spatial Description}}: Describes the placing motion relative to the entire workspace, which is divided into a $3\times3$ grid (e.g., ``top left,'' ``center''). This is used when the placement location is not near any other object. For example: ``Pick the yellow block and place it in the middle left of the workspace.''
    \item \textbf{\textit{Relative Spatial Description}}: Describes the placing motion relative to a nearby reference object. This is used when the placement is close to another object. For example: ``Pick the yellow block and place it to the top right of the mustard bottle.''
\end{itemize}

Humans naturally switch between these description styles. To mimic this, we generate descriptions based on the normalized distance between the placing location and the closest object, as shown in Fig.~\ref{fig:exman_thresholding}. If the distance is less than $d_{\mathit{rel}}$, we generate a relative description. If it is greater than $d_{\mathit{abs}}$, we generate an absolute one. 
In our experiments, we use $d_{\mathit{abs}}=0.15$ and $d_{\mathit{rel}}=0.3$ as we empirically found that these values tend to generate a balanced distribution between relative and absolute descriptions.

\subsection{Unified Model via Two-Stage Fine-Tuning}
A primary challenge in robot learning is the scarcity of robot-specific demonstration data compared to the abundance of general computer vision datasets. To address this, we propose a data-efficient, two-stage fine-tuning strategy that leverages a Chain-of-Thought (CoT) \cite{wei2022chain} reasoning process.

\textbf{Stage 1: Object Grounding Pre-training.} We first pre-train our VLM backbone on an object grounding task. Using a dataset of 8,000 images with object labels and object center locations, we teach the model to perform a foundational vision task: identifying all objects in an image and listing their names and center coordinates. This task involves outputting a set of detected objects $\objset = \{(n_i, p_i)\}_{i=1}^N$, where $n_i$ is the object name and $p_i$ is the center coordinate for the detected object $i$. This pre-training endows the model with a robust visual understanding that serves as a strong prior for downstream manipulation tasks.

\textbf{Stage 2: CoT-based Multi-task Fine-tuning.} We then fine-tune the model on our smaller, robot-specific dataset of 1,000 demonstrations for the three synergistic tasks of LACY. Fig.~\ref{fig:overview} shows the input and output format for each model: $\LtoA$, $\AtoL$, and $\LtoC$.
Crucially, all three tasks are formulated as a CoT process that explicitly utilizes the pre-trained grounding skill. The model is prompted to first perform the object grounding task to predict object name and location, $\hat{\objset}$, and then use that as context to perform $\LtoA$, $\AtoL$, and $\LtoC$. This CoT approach not only makes the model's reasoning process more transparent and robust but also allows the grounding skill to be further refined with in-domain robotics data. This two-stage approach allows LACY to achieve high performance with significantly less robot-specific data. Throughout this process, we use Low-Rank Adaptation (LoRA)~\cite{hu2021lora} for efficient fine-tuning.

\subsection{Self-Improving Data Generation with $\LtoA$ and $\AtoL$}
The central mechanism of LACY is its ability to generate its own training data through a closed loop, which we term the $\LtoAtoL$ pipeline. This process begins with an input language command $\lan$ from our training set, which is fed into $\LtoA$ to generate an intermediate action $\hat{\act}$. This generated action $\hat{\act}$ is then immediately fed into $\AtoL$, along with the same observation $\obs$, to produce a new, reconstructed language description $\hat{\lan}$. This cycle yields a new complete triplet, $\demo_{new}=(\obs, \lan, \hat{\act})$, which can be added to our training set if its quality and semantic consistency are verified by the $\LtoC$ module.


\subsection{Confidence-based Data Augmentation with $\LtoC$}
\label{sec:data_sampling}
A naive approach to data augmentation would be to generate many samples and filter them. However, this can lead to overfitting on tasks the model already performs well on, and could add redundant data if multiple outputs from a single sample are deemed valid. To address this, we propose a confidence-based active data augmentation strategy guided by $\LtoC$ that incorporates a robust, voting-based filtering mechanism, as detailed in Algorithm~\ref{alg:checkman}.

\begin{figure}[t]
    \begin{center}
        \includegraphics[width=1.0\columnwidth]{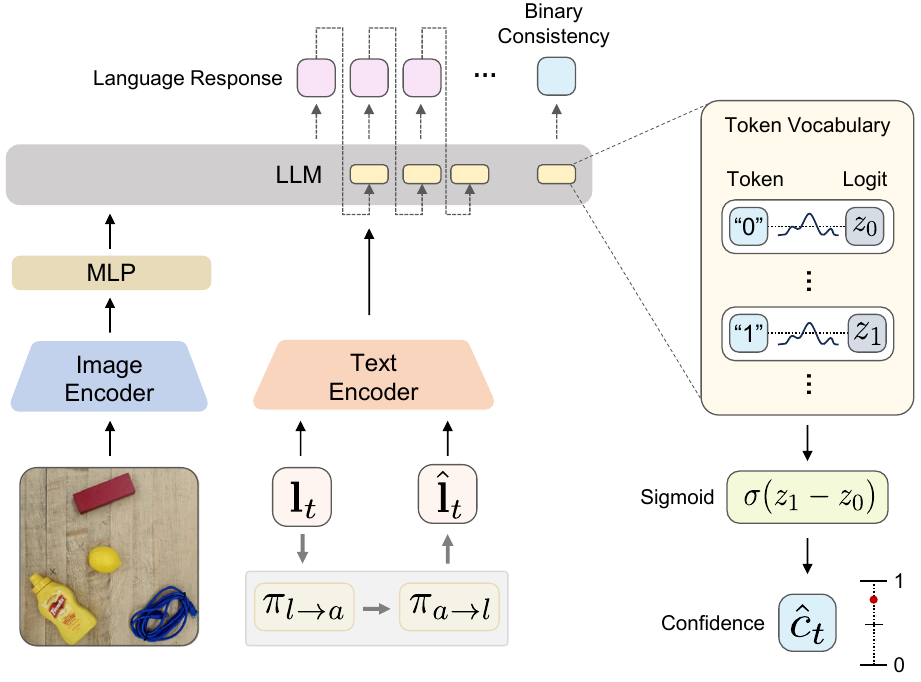}%
    \end{center}
    \vspace{-8pt}
    \caption{\textbf{Binary confidence extraction from VLM outputs.} The logits $z_0$ and $z_1$ corresponding to the tokens ``0'' and ``1'' are used to compute a confidence score $c$.}
    \label{fig:consistency_mechanism}
\end{figure}

The process begins with a single, deterministic pass through the $\LtoAtoL$ pipeline. For a given input language instruction $\lan_t$, we obtain a generated action $\hat{\act}_t$ and a reconstructed language description $\hat{\lan}_t$. Instead of having the VLM directly output a numerical value in text form, which can be unreliable~\cite{manggala2024qa}, we task $\LtoC$ with a binary classification problem. It outputs a consistency value "1" if $\lan_t$ and $\hat{\lan}_t$ are consistent, and "0" otherwise. 
To obtain a confidence value for binary classification, we directly extract the logits corresponding to the tokens "0" and "1" from the model output at the final decoding step, as shown in Fig.~\ref{fig:consistency_mechanism}. Instead of selecting the discrete token, we apply a sigmoid function to the logit difference to obtain a continuous probability in the range of $[0,1]$.

This technique of using logit differences to calibrate model confidence is inspired by similar methods in language and vision models~\cite{kadavath2022language, radford2021learning, shah2024lm}. The continuous consistency score $c$ is calculated as a probability:
\begin{equation}
c = \sigma(z_1 - z_0)
\end{equation}
where $\sigma(\cdot)$ is the sigmoid function. Note that this formulation is equivalent to the two-class softmax:
\begin{equation}
\sigma(z_1 - z_0) = \frac{\exp(z_1)}{\exp(z_0) + \exp(z_1)}.
\end{equation}

\begin{itemize}
\item If the consistency score $c$ is high (i.e., $c \geq \tau$), we consider this a high-confidence case that the model has already mastered. No additional data are generated for this sample, avoiding redundant computation.
\end{itemize}

For each candidate action $\act'_{i} \in \mathcal{A}_{\text{cand}}$, we then generate a corresponding group of $N$ language descriptions ($\mathcal{L}_{\text{cand}}$) using $\AtoL$. Instead of accepting an action upon the first successful validation, we evaluate the entire group. An action $\hat{\act}_i$ is deemed valid and added to our new dataset only if a majority ($\geq\nu$) of its corresponding language descriptions are judged as consistent with the original instruction $\lan_t$ by $\LtoC$. This majority-voting approach prevents the addition of duplicate ($\lan, \hat{\act}$) pairs and ensures that only actions that are robustly and consistently explainable are used for retraining, filtering out potential flukes.

Finally, to prevent catastrophic forgetting, we re-initialize the model from the base model on a merged dataset including both original data and newly generated, high-quality data.

\begin{algorithm}[t]
\caption{Confidence-based Active Data Augmentation with Majority Voting}\label{alg:checkman}
\KwIn{Initial demonstration set $\demoset$, LACY model $\policy$, number of stochastic samples $N$, consistency threshold $\tau$, voting threshold $\nu=0.5$}
\KwOut{Augmented dataset $\demoset_{\text{aug}}$}
$\demoset_{\text{new}} \leftarrow \emptyset$\;
\For{each $(\obs_t, \lan_{t}, \act_t) \in \demoset$}
{
$\hat{\act}_t \leftarrow \LtoA(\obs_t, \lan_t)$\;
$\hat{\lan}_t \leftarrow \AtoL(\obs_t, \hat{\act}_t)$\;
$c \leftarrow \LtoC(\obs_t, \lan_t, \hat{\lan}_t)$\;
\If{$c \leq \tau$}{
$\mathcal{A}_{\text{cand}} \leftarrow \text{Stochastic\LtoA}(\obs_t, \lan_t, N)$\;
\For{each $\hat{\act}_i \in \mathcal{A}_{\text{cand}}$}{
$N_p \leftarrow 0$\;
$\mathcal{L}_{\text{cand}} \leftarrow \text{Stochastic\AtoL}(\obs_t, \hat{\act}_i, N)$\;
\For{each $\hat{\lan}_{j} \in \mathcal{L}_{\text{cand}}$}{
$c \leftarrow \LtoC(\obs_t, \lan_t, \hat{\lan}_{j})$\;
\If{$c \geq \tau$}{
$N_p \leftarrow N_p + 1$\;
}
}
\If{$N_p / N \geq \nu$}{
$\demoset_{\text{new}} \leftarrow \demoset_{\text{new}} \cup \{ \obs_t, \lan_t, \hat{\act}_i \}$
}
}
}
$\demoset_{\text{aug}} \leftarrow \demoset \cup \demoset_{\text{new}}$\;}
\end{algorithm}

\section{Experiments}
\subsection{Experimental Setup}

\begin{figure}[t]
    \begin{center}
        \includegraphics[width=\columnwidth]{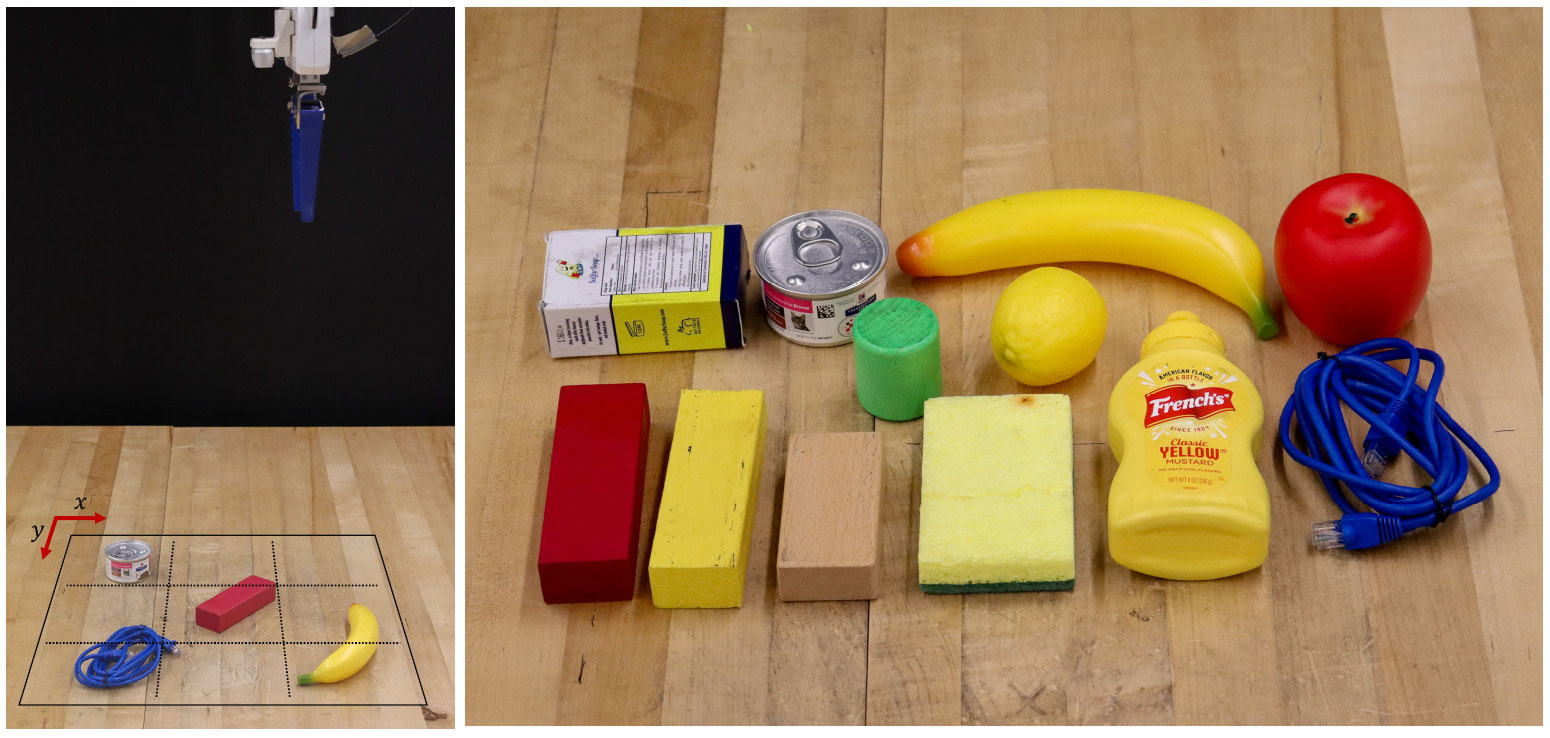}%
    \end{center}
    \vspace{-10pt}
    \caption{\textbf{Real robot experiment setup.} (Left) The workspace is divided into a $3\times3$ grid to provide an absolute spatial reference for task descriptions. A top-view image captured by an Intel RealSense D415 camera serves as the visual input to LACY. (Right) Objects used in the real-robot experiment, including both everyday items and selected YCB objects.}
    \label{fig:experiment_setup}
\end{figure}

Our experiments are conducted in a simulated tabletop environment built with CoppeliaSim~\cite{coppeliaSim} and 32 YCB objects, and in a real-world setup with a Franka Emika Panda Robot and an Intel RealSense D415 camera using 12 real-world objects, as shown in Fig.~\ref{fig:experiment_setup}.

\textbf{Dataset:} We use a training dataset comprising up to 4,000 successful pick-and-place demonstrations. For ablation studies, we employ a smaller subset of 1,000 demonstrations to evaluate the effectiveness of our proposed methods under data-scarce conditions, while the simulation test set includes 100 unseen scenarios. For the real-world environment, we trained the model with 212 demonstrations and tested with 50 unseen scenarios.

\textbf{Implementation Details:} The object grounding pre-training is conducted for 2 epochs. The second stage of the multi-task fine-tuning is performed for 5 epochs, as well as the multi-task re-fine-tuning using the augmented dataset including both original data and data sampled via the $\LtoC$ module. All experiments are carried out on a single NVIDIA A40 GPU with 48 GB of memory.

\textbf{Evaluation Metrics:} We evaluate our framework using the following metrics:
\begin{itemize}
    \item \textbf{L2A (\%):} 
    The task success rate of the $\LtoA$ module. A task is considered successful if the correct object is grasped and placed at the target location which semantically meets the given spatial description.
    \item \textbf{A2L (\%):} 
    The task success rate of the $\AtoL$ module. A task is considered successful if it correctly mentions the target object to grasp and accurately describes the spatial relationship for placement given the action $\act_t$.
    
    \item \textbf{L2C (\%):} The accuracy of the $\LtoC$ module in detecting semantically consistent language-action pairs.
    \item \textbf{Pick / Pick \& Place (\%):} In real-world experiments, the percentage of successful grasping and placement completed by the robot following model-generated actions.
\end{itemize}

\begin{table}
\caption{Reasoning Capability Results in Simulation}
\label{tab:sim_reasoning}
\centering
\begin{tabular}{@{}lcccc@{}}
\toprule
{\textbf{Model}} & L2A (\%) & A2L (\%) & L2C (\%) \\ \midrule
GPT-4o w/ Grounding  & \underline{90} & 39 & 8 \\
GPT-4o w/o Grounding & 28 & \underline{40} & \underline{76} \\
LLaVA-NeXT (base)    & 6 & 6 & 50 \\
LACY (4k data, ours) & \textbf{95} & \textbf{76} & \textbf{95} \\ \bottomrule
\end{tabular}
\end{table}

\subsection{Reasoning Capabilities in Simulation}
First, we evaluate the core reasoning capabilities of our bidirectional model. To establish a strong performance baseline, we use a version of our LACY model fine-tuned on a larger dataset of 4,000 demonstrations. We compare it against several baselines to justify the need for fine-tuning a specialized model. The results are presented in Table~\ref{tab:sim_reasoning}.

\textbf{Baselines:} We compare our \textit{LACY-Joint (4k data)} model with \textit{LLaVA-NeXT (base)}, which has not been fine-tuned on robotics data, and \textit{GPT-4o}, both with and without ground-truth object location information provided in the prompt.

\textbf{Analysis:} The results highlight that large, general-purpose models like GPT-4o and the base LLaVA-NeXT model struggle with precise spatial grounding from a single image, leading to poor performance. However, when provided with explicit grounding information, GPT-4o achieves near-perfect results, demonstrating its powerful reasoning capability when ambiguity in object grounding is removed. Our fine-tuned LACY model significantly outperforms the non-fine-tuned baselines, showing the effectiveness of our training approach.

\subsection{Ablation Studies}
We conduct a series of ablation studies to validate each component of our proposed framework. As our goal is to improve performance in data-limited scenarios, all models in these studies are trained on a reduced dataset of 1,000 demonstrations.

\begin{table}[h]
\caption{Ablation Study on Joint Training and Filtering}
\label{tab:ablation_filtering}
\centering
\begin{tabular}{@{}lccc@{}}
\toprule
\textbf{Model} & L2A (\%) & A2L (\%) & L2C (\%) \\ \midrule
LACY-Ind & 78 & 78 & \underline{92} \\
LACY-Joint (ours) & \underline{83} & \underline{80} & \underline{92} \\
LACY-Joint-SI (ours) & \textbf{93} & \textbf{85} & \textbf{95} \\ \bottomrule
\end{tabular}
\end{table}

\textbf{Effectiveness of Joint Training and Filtering:} We first investigate the benefits of our joint training approach and the self-improvement filtering mechanism. As shown in Table~\ref{tab:ablation_filtering}, the jointly trained model (\textit{LACY-Joint}) outperforms the independently trained one (\textit{LACY-Ind}), supporting our hypothesis that a shared representation through joint training is beneficial. Furthermore, adding our self-improvement cycle (\textit{LACY-Joint-SI}) provides a substantial performance boost across all metrics, confirming the effectiveness of our self-improvement pipeline.

\textbf{Necessity of CoT:} We examine the impact of our Chain-of-Thought (CoT) prompting strategy. Table~\ref{tab:ablation_cot} compares our CoT-based model with a version that directly generates the output without the intermediate object grounding step (\textit{LACY-non-CoT}). The results indicate that the explicit reasoning step provided by CoT is crucial for achieving higher performance.

\begin{table}[h]
\caption{Ablation Study on Chain-of-Thought Prompting}
\label{tab:ablation_cot}
\centering
\begin{tabular}{@{}lcccc@{}}
\toprule
\textbf{Model}  & L2A (\%) & A2L (\%) & L2C (\%) \\ \midrule
LACY-non-CoT    & 52             & 43             & 84             \\
LACY-CoT (ours) & \textbf{83}               & \textbf{80}             & \textbf{92}              \\ \bottomrule
\end{tabular}
\end{table}



\begin{figure*}[t]
    \begin{center}
        \includegraphics[width=1.0\textwidth]{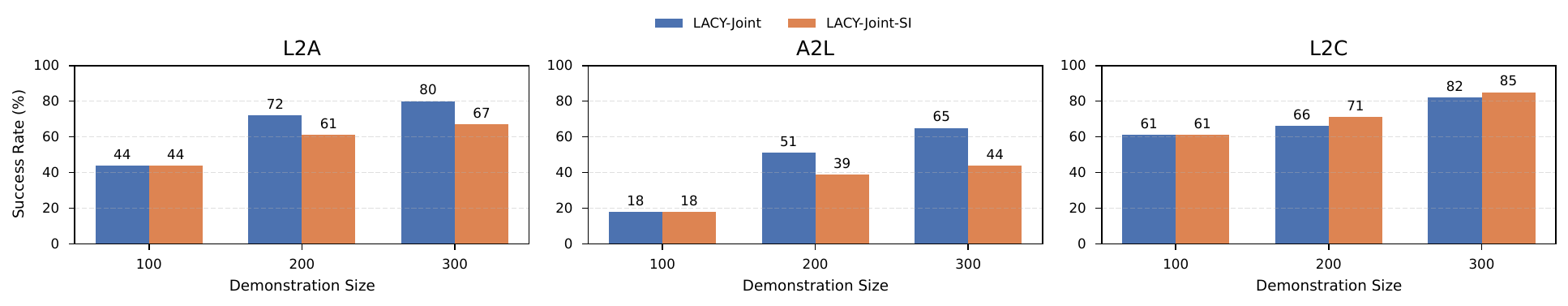}%
    \end{center}
    \vspace{-10pt}
    \caption{\textbf{Self-improvement of LACY.} LACY-Joint is trained only on ground-truth data, while LACY-Joint-SI is trained on ground-truth plus $\LtoC$-sampled data.}
    \label{fig:self_improvement}
\end{figure*}

\subsection{Self-Improvement Capability}
To evaluate the model's ability to improve through self-generated data, we assess its performance after varying numbers of self-improvement iterations (from one to three), starting from the model trained on only 100 demonstration data points. In each iteration, the model performs data augmentation via the $\LtoC$ task and fine-tuning with the combined dataset (Algorithm~\ref{alg:checkman}), generating 100 new triplets $(\obs, \lan, \hat{\act})$. Thus, a model with $n$ self-improvement iterations is trained on $100\times(n+1)$ data points. Fig.~\ref{fig:self_improvement} reports the task success rates for models trained on the corresponding dataset sizes. 
    \textit{LACY-Joint} is trained solely on the ground-truth dataset, while \textit{LACY-Joint-SI} is trained on a mixture of the ground-truth data and the additional data generated through the $\LtoC$-based data augmentation process. While \textit{LACY-Joint-SI} initially underperforms \textit{LACY-Joint} on $\LtoA$ and $\AtoL$, its success rates increase consistently as more self-generated data is incorporated, and it notably surpasses \textit{LACY-Joint} on $\LtoC$.


\subsection{Real-World Evaluation}

\begin{figure}[t]
    \centering
    \includegraphics[width=1.0\columnwidth]{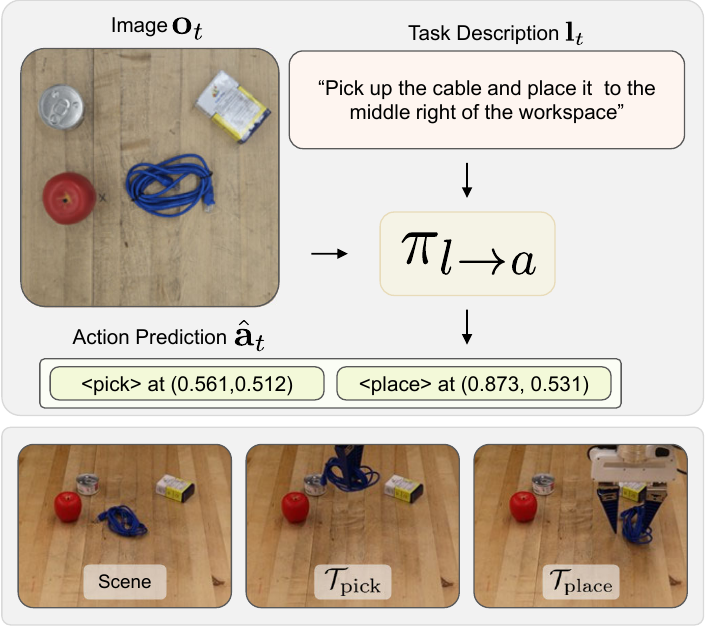}
    \vspace{-10pt}
    \caption{\textbf{Real Robot Reasoning.} (Top) Given an image observation $\obs_t$ and a task description $\lan_t$, the robot reasons the appropriate pick-and-place action $\hat{\act}_t$ via $\LtoA$. (Bottom) The robot grasps the cable and places it in the designated location.}
    \label{fig:realrobot}
\end{figure}
Finally, we evaluate LACY on real data and compare the model performances on a 7-DoF Franka Emika Panda robot. Refer to the real robot experiment setup in Fig.~\ref{fig:realrobot}.
We evaluate the model's ability to generate correct actions, language, and consistency decisions based on real-world images, and separately report the physical grasp success rate. 
To investigate the sim-to-real gap, we compare the performances between the models by ablating training with real-world data. \textit{LACY-Ind-Real} and \textit{LACY-Joint-Real} are further trained on a real-world dataset from the simulation-trained models \textit{LACY-Ind} and \textit{LACY-Joint}. The results in Table~\ref{tab:real_world_image_eval} show that the performance gains observed in simulation translate effectively to the real world. Notably, both \textit{LACY-Ind} and \textit{LACY-Joint} successfully detect most of the objects in the given image, and even accurately locate the objects that are both in the simulation and real-world (e.g., mustard bottle, sponge). Most of the failure cases are due to misnaming the objects in the image. Through training with real-world data, \textit{LACY-Ind-Real} and \textit{LACY-Joint-Real} improve the performance by successfully deriving correct object names, with \textit{LACY-Joint-Real} accomplishing the best performance across all three tasks. Table~\ref{tab:real_world_eval} shows that the pick-and-place actions generated by the LACY models are also effective in real-robot tasks. 

\begin{table}[h]
\caption{Real-World Data Evaluation}
\label{tab:real_world_image_eval}
\centering
\begin{tabular}{@{}lcccc@{}}
\toprule
\textbf{Model}     & L2A (\%) & A2L (\%) & L2C (\%) \\ \midrule
LACY-Ind      & 78               & 36             & \underline{94} \\
LACY-Ind-Real     & \underline{82}               & \underline{80}             & \underline{94} \\
LACY-Joint    & 80               & 28             & \textbf{98} \\
LACY-Joint-Real    & \textbf{88}               & \textbf{88}             & \textbf{98}                 \\ \bottomrule
\end{tabular}
\end{table}

\begin{table}[h]
\caption{Real-World Robot Evaluation}
\label{tab:real_world_eval}
\centering
\begin{tabular}{@{}lcc@{}}
\toprule
\textbf{Model} & Pick (\%) & Pick \& Place (\%)\\ \midrule
LACY-Ind-Real & 65 & 55 \\
LACY-Joint-Real & \textbf{72.5} & \textbf{60}  \\ \bottomrule
\end{tabular}
\end{table}

\section{Conclusion}
This paper presented LACY, a VLM-based framework that achieves bidirectional grounding between language and action for robotic manipulation. By unifying action generation ($\LtoA$), action explanation ($\AtoL$), and semantic verification ($\LtoC$) within a single model, LACY enables a novel self-improving cycle that autonomously generates and filters its own training data. Experimental results demonstrate that the synergy between joint bidirectional training and confidence-based active data augmentation significantly improves manipulation success rates in both simulation and real-world settings. These findings highlight that moving beyond the prevailing unidirectional $\LtoA$ paradigm, and equipping agents with the ability to both act and explain offers a promising path toward more robust, data-efficient, and generalizable robotic systems. 

Despite these encouraging results, several limitations remain and warrant further investigation. Notably, the $\LtoC$ semantic verification module is not specifically trained to evaluate object grounding quality, making the data generation pipeline susceptible to errors from incorrect object grounding outputs. Because object grounding results play a critical role in the model’s behavior through CoT-based fine-tuning, such errors may propagate downstream, leading the model to misidentify picking locations and ultimately reducing task success rates. Future work will address these limitations while expanding the framework’s capabilities. In particular, we plan to (1) develop more robust perception and verification modules that can better handle detection uncertainties and (2) extend the LACY framework to more complex, long-horizon tasks and a broader range of manipulation skills. We believe this bidirectional perspective can serve as a foundation for scalable robot intelligence.

\vspace{-2pt}
{
\bibliography{references} 

@inproceedings{jang2022bc,
  title={Bc-z: Zero-shot task generalization with robotic imitation learning},
  author={Jang, Eric and Irpan, Alex and Khansari, Mohi and Kappler, Daniel and Ebert, Frederik and Lynch, Corey and Levine, Sergey and Finn, Chelsea},
  booktitle={Conference on Robot Learning},
  pages={991--1002},
  year={2022},
  organization={PMLR}
}

@article{zhao2023learning,
  title={Learning fine-grained bimanual manipulation with low-cost hardware},
  author={Zhao, Tony Z and Kumar, Vikash and Levine, Sergey and Finn, Chelsea},
  journal={arXiv preprint arXiv:2304.13705},
  year={2023}
}

@article{chi2023diffusion,
  title={Diffusion policy: Visuomotor policy learning via action diffusion},
  author={Chi, Cheng and Xu, Zhenjia and Feng, Siyuan and Cousineau, Eric and Du, Yilun and Burchfiel, Benjamin and Tedrake, Russ and Song, Shuran},
  journal={The International Journal of Robotics Research},
  pages={02783649241273668},
  year={2023},
  publisher={SAGE Publications Sage UK: London, England}
}

@article{ze20243d,
  title={3d diffusion policy: Generalizable visuomotor policy learning via simple 3d representations},
  author={Ze, Yanjie and Zhang, Gu and Zhang, Kangning and Hu, Chenyuan and Wang, Muhan and Xu, Huazhe},
  journal={arXiv preprint arXiv:2403.03954},
  year={2024}
}

@inproceedings{radford2021learning,
  title={Learning transferable visual models from natural language supervision},
  author={Radford, Alec and Kim, Jong Wook and Hallacy, Chris and Ramesh, Aditya and Goh, Gabriel and Agarwal, Sandhini and Sastry, Girish and Askell, Amanda and Mishkin, Pamela and Clark, Jack and others},
  booktitle={International conference on machine learning},
  pages={8748--8763},
  year={2021},
  organization={PmLR}
}

@inproceedings{zhai2023sigmoid,
  title={Sigmoid loss for language image pre-training},
  author={Zhai, Xiaohua and Mustafa, Basil and Kolesnikov, Alexander and Beyer, Lucas},
  booktitle={Proceedings of the IEEE/CVF international conference on computer vision},
  pages={11975--11986},
  year={2023}
}

@inproceedings{li2023blip,
  title={Blip-2: Bootstrapping language-image pre-training with frozen image encoders and large language models},
  author={Li, Junnan and Li, Dongxu and Savarese, Silvio and Hoi, Steven},
  booktitle={International conference on machine learning},
  pages={19730--19742},
  year={2023},
  organization={PMLR}
}

@article{liu2023visual,
  title={Visual instruction tuning},
  author={Liu, Haotian and Li, Chunyuan and Wu, Qingyang and Lee, Yong Jae},
  journal={Advances in neural information processing systems},
  volume={36},
  pages={34892--34916},
  year={2023}
}

@misc{liu2024llavanext,
    title={LLaVA-NeXT: Improved reasoning, OCR, and world knowledge},
    url={https://llava-vl.github.io/blog/2024-01-30-llava-next/},
    author={Liu, Haotian and Li, Chunyuan and Li, Yuheng and Li, Bo and Zhang, Yuanhan and Shen, Sheng and Lee, Yong Jae},
    month={January},
    year={2024}
}

@inproceedings{shridhar2023perceiver,
  title={Perceiver-actor: A multi-task transformer for robotic manipulation},
  author={Shridhar, Mohit and Manuelli, Lucas and Fox, Dieter},
  booktitle={Conference on Robot Learning},
  pages={785--799},
  year={2023},
  organization={PMLR}
}

@article{yang2022interactive,
  title={Interactive robotic grasping with attribute-guided disambiguation},
  author={Yang, Yang and Lou, Xiaoman and Choi, Changhyun},
  journal={IEEE Robotics and Automation Letters},
  volume={7},
  number={2},
  pages={4439--4446},
  year={2022},
  publisher={IEEE}
}

@misc{brohan2023rt2,
      title={RT-2: Vision-Language-Action Models Transfer Web Knowledge to Robotic Control}, 
      author={Anthony Brohan and Noah Brown and Justice Carbajal and Yevgen Chebotar and Xi Chen and Krzysztof Choromanski and Tianli Ding and Danny Driess and Avinava Dubey and Chelsea Finn and Pete Florence and Chuyuan Fu and Montse Gonzalez Arenas and Keerthana Gopalakrishnan and Kehang Han and Karol Hausman and Alexander Herzog and Jasmine Hsu and Brian Ichter and Alex Irpan and Nikhil Joshi and Ryan Julian and Dmitry Kalashnikov and Yuheng Kuang and Isabel Leal and Lisa Lee and Tsang-Wei Edward Lee and Sergey Levine and Yao Lu and Henryk Michalewski and Igor Mordatch and Karl Pertsch and Kanishka Rao and Krista Reymann and Michael Ryoo and Grecia Salazar and Pannag Sanketi and Pierre Sermanet and Jaspiar Singh and Anikait Singh and Radu Soricut and Huong Tran and Vincent Vanhoucke and Quan Vuong and Ayzaan Wahid and Stefan Welker and Paul Wohlhart and Jialin Wu and Fei Xia and Ted Xiao and Peng Xu and Sichun Xu and Tianhe Yu and Brianna Zitkovich},
      year={2023},
      eprint={2307.15818},
      archivePrefix={arXiv},
      primaryClass={cs.RO},
      url={https://arxiv.org/abs/2307.15818}, 
}

@misc{shridhar2022cliport,
      title={CLIPort: What and Where Pathways for Robotic Manipulation}, 
      author={Mohit Shridhar and Lucas Manuelli and Dieter Fox},
      year={2021},
      eprint={2109.12098},
      archivePrefix={arXiv},
      primaryClass={cs.RO},
      url={https://arxiv.org/abs/2109.12098}, 
}

@article{kim2024openvla,
  title={OpenVLA: An Open-Source Vision-Language-Action Model},
  author={Kim, Moo Jin and Pertsch, Karl and Karamcheti, Siddharth and Xiao, Ted and Balakrishna, Ashwin and Nair, Suraj and Rafailov, Rafael and Foster, Ethan and Lam, Grace and Sanketi, Pannag and others},
  journal={arXiv preprint arXiv:2406.09246},
  year={2024}
}

@article{pari2024finetuning,
  title={Fine-Tuning Vision-Language-Action Models: Optimizing Speed and Success},
  author={Pari, Aniketh and Black, Kevin and Xu, Charles and Walke, Homer and Dasari, Sudeep and Kumar, Avi and Rajeswaran, Aravind and Finn, Chelsea and Levine, Sergey},
  journal={arXiv preprint arXiv:2405.08232},
  year={2024}
}

@article{pertsch2025fast,
  title={Fast: Efficient action tokenization for vision-language-action models},
  author={Pertsch, Karl and Stachowicz, Kyle and Ichter, Brian and Driess, Danny and Nair, Suraj and Vuong, Quan and Mees, Oier and Finn, Chelsea and Levine, Sergey},
  journal={arXiv preprint arXiv:2501.09747},
  year={2025}
}

@inproceedings{walke2023bridgedata,
  title={Bridgedata v2: A dataset for robot learning at scale},
  author={Walke, Homer Rich and Black, Kevin and Zhao, Tony Z and Vuong, Quan and Zheng, Chongyi and Hansen-Estruch, Philippe and He, Andre Wang and Myers, Vivek and Kim, Moo Jin and Du, Max and others},
  booktitle={Conference on Robot Learning},
  pages={1723--1736},
  year={2023},
  organization={PMLR}
}

@article{liu2023libero,
  title={Libero: Benchmarking knowledge transfer for lifelong robot learning},
  author={Liu, Bo and Zhu, Yifeng and Gao, Chongkai and Feng, Yihao and Liu, Qiang and Zhu, Yuke and Stone, Peter},
  journal={Advances in Neural Information Processing Systems},
  volume={36},
  pages={44776--44791},
  year={2023}
}

@article{qian2024thinkgrasp,
  title={ThinkGrasp: A vision-language system for strategic part grasping in clutter},
  author={Qian, Yaoyao and Zhu, Xupeng and Biza, Ondrej and Jiang, Shuo and Zhao, Linfeng and Huang, Haojie and Qi, Yu and Platt, Robert},
  journal={arXiv preprint arXiv:2407.11298},
  year={2024}
}

@misc{li2025hamster,
      title={HAMSTER: Hierarchical Action Models For Open-World Robot Manipulation}, 
      author={Yi Li and Yuquan Deng and Jesse Zhang and Joel Jang and Marius Memmel and Raymond Yu and Caelan Reed Garrett and Fabio Ramos and Dieter Fox and Anqi Li and Abhishek Gupta and Ankit Goyal},
      year={2025},
      eprint={2502.05485},
      archivePrefix={arXiv},
      primaryClass={cs.RO},
      url={https://arxiv.org/abs/2502.05485}, 
}

@article{yuan2024robopoint,
  title={Robopoint: A vision-language model for spatial affordance prediction for robotics},
  author={Yuan, Wentao and Duan, Jiafei and Blukis, Valts and Pumacay, Wilbert and Krishna, Ranjay and Murali, Adithyavairavan and Mousavian, Arsalan and Fox, Dieter},
  journal={arXiv preprint arXiv:2406.10721},
  year={2024}
}

@misc{nasiriany2024,
      title={RT-Affordance: Affordances are Versatile Intermediate Representations for Robot Manipulation}, 
      author={Soroush Nasiriany and Sean Kirmani and Tianli Ding and Laura Smith and Yuke Zhu and Danny Driess and Dorsa Sadigh and Ted Xiao},
      year={2024},
      eprint={2411.02704},
      archivePrefix={arXiv},
      primaryClass={cs.RO},
      url={https://arxiv.org/abs/2411.02704}, 
}

@inproceedings{yu2025parameter,
  title={A parameter-efficient tuning framework for language-guided object grounding and robot grasping},
  author={Yu, Houjian and Li, Mingen and Rezazadeh, Alireza and Yang, Yang and Choi, Changhyun},
  booktitle={2025 IEEE International Conference on Robotics and Automation (ICRA)},
  pages={14353--14360},
  year={2025},
  organization={IEEE}
}

@inproceedings{hu2021lora,
  title={LoRA: Low-Rank Adaptation of Large Language Models},
  author={Hu, Edward J and Shen, Yelong and Wallis, Phillip and Allen-Zhu, Zeyuan and Li, Yuanzhi and Wang, Shean and Wang, Lu and Chen, Weizhu},
  booktitle={International Conference on Learning Representations},
  year={2021}
}

@inproceedings{mandlekar2023mimicgen,
  title={Mimicgen: A data generation system for scalable robot learning using human demonstrations},
  author={Mandlekar, Ajay and Nasiriany, Soroush and Wen, Bowen and Akinola, Iretiayo and Narang, Yashraj and Fan, Linxi and Zhu, Yuke and Fox, Dieter},
  booktitle={Conference on Robot Learning},
  pages={1820--1864},
  year={2023},
  organization={PMLR}
}

@inproceedings{kalashnikov2018scalable,
  title={Scalable Deep Reinforcement Learning for Vision-Based Robotic Manipulation},
  author={Kalashnikov, Dmitry and Irpan, Alex and Pastor, Peter and others},
  booktitle={Conference on Robot Learning},
  year={2018}
}

@inproceedings{kumar2021rma,
  title={Rapid Motor Adaptation for Legged Robots},
  author={Kumar, Ashish and Fu, Zipeng and Pathak, Deepak and others},
  booktitle={Robotics: Science and Systems},
  year={2021}
}

@inproceedings{burda2018exploration,
  title={Exploration by Random Network Distillation},
  author={Burda, Yuri and Edwards, Harrison and Storkey, Amos and others},
  booktitle={International Conference on Learning Representations},
  year={2018}
}

@inproceedings{srinivas2020curl,
  title={CURL: Contrastive Unsupervised Representations for Reinforcement Learning},
  author={Srinivas, Aravind and Laskin, Michael and Abbeel, Pieter},
  booktitle={International Conference on Machine Learning},
  year={2020}
}

@inproceedings{kostrikov2020image,
  title={Image Augmentation Is All You Need: Regularizing Deep Reinforcement Learning from Pixels},
  author={Kostrikov, Ilya and Yarats, Denis and Fergus, Rob},
  booktitle={International Conference on Learning Representations},
  year={2020}
}

@inproceedings{sermanet2018time,
  title={Time-Contrastive Networks: Self-Supervised Learning from Video},
  author={Sermanet, Pierre and Lynch, Corey and Chebotar, Yevgen and others},
  booktitle={International Conference on Robotics and Automation},
  year={2018}
}

@article{bai2023qwen,
  title={Qwen-vl: A versatile vision-language model for understanding, localization, text reading, and more},
  author={Bai, Jinze and Bai, Shuai and Yang, Shusheng and Wang, Shijie and Tan, Sinan and Wang, Peng and Lin, Junyang and Zhou, Chang and Zhou, Jingren},
  journal={arXiv preprint arXiv:2308.12966},
  year={2023}
}

@article{kadavath2022language,
  title={Language models (mostly) know what they know},
  author={Kadavath, Saurav and Conerly, Tom and Askell, Amanda and Henighan, Tom and Drain, Dawn and Perez, Ethan and Schiefer, Nicholas and Jones, Anna and Joseph, Nicholas and DasSarma, Nova and others},
  journal={arXiv preprint arXiv:2207.05221},
  year={2022}
}

@inproceedings{shah2024lm,
  title={{LM-Nav}: Robotic Navigation with Large Language Models},
  author={Shah, Dhruv and Liang, Michael and Liu, Yunchu and Naren, Ajay S and Stone, Garrett and Kumar, Ajay and Scherer, Sebastian and Gupta, Abhinav},
  booktitle={Proceedings of the IEEE/CVF Conference on Computer Vision and Pattern Recognition},
  pages={18052--18062},
  year={2024}
}

@inproceedings{zeng2021transporter,
  title={Transporter networks: Rearranging the visual world for robotic manipulation},
  author={Zeng, Andy and Florence, Phillip and Tompson, Jonathan and Welker, Stefan and Chien, Jonathan and Attarian, Maria and Armstrong, Travis and Krasin, Ivan and Duong, Dan and Sindhwani, Vikas and others},
  booktitle={Conference on Robot Learning},
  pages={103--120},
  year={2021},
  organization={PMLR}
}

@article{huang2023large,
  title={Large language models cannot self-correct reasoning yet},
  author={Huang, Jie and Gu, Xinyun and Hou, Linyi and Wang, Jiawei and Li, Jiacheng and Chen, Granz and Chen, Chaoyun and Liu, Zhen and Zhang, Yue and Gui, Tao and others},
  journal={arXiv preprint arXiv:2310.01798},
  year={2023}
}

@article{valmeekam2023can,
  title={Can llms really grasp simple causal structures?},
  author={Valmeekam, K and Marquez, M and Kumar, S and Sridharan, M and Kambhampati, S},
  journal={arXiv preprint arXiv:2305.15769},
  year={2023}
}

@inproceedings{zhu2017unpaired,
  title={Unpaired image-to-image translation using cycle-consistent adversarial networks},
  author={Zhu, Jun-Yan and Park, Taesung and Isola, Phillip and Efros, Alexei A},
  booktitle={Proceedings of the IEEE international conference on computer vision},
  pages={2223--2232},
  year={2017}
}

@inproceedings{pmlr-v164-ajay22a,
  title = 	 {Cycle-Consistent Inverse Dynamics for Visual Imitation Learning},
  author =       {Ajay, Anurag and Saber, Yevgen and Roh, B and Jaakkola, T},
  booktitle = 	 {Proceedings of the Thirty-First International Joint Conference on Artificial Intelligence},
  pages = 	 {4359--4366},
  year = 	 {2022},
  editor = 	 {L. De Raedt},
  publisher =    {International Joint Conferences on Artificial Intelligence Organization},
}

@article{wei2022chain,
  title={Chain-of-thought prompting elicits reasoning in large language models},
  author={Wei, Jason and Wang, Xuezhi and Schuurmans, Dale and Bosma, Maarten and Xia, Fei and Chi, Ed and Le, Quoc V and Zhou, Denny and others},
  journal={Advances in neural information processing systems},
  volume={35},
  pages={24824--24837},
  year={2022}
}

@inproceedings{coppeliaSim,
    author="E. Rohmer and S. P. N. Singh and M. Freese",
    title="CoppeliaSim (formerly V-REP): a Versatile and
        Scalable Robot Simulation Framework",
    booktitle="Proc. of The International Conference on
        Intelligent Robots and Systems (IROS)",
    year="2013",
    note="www.coppeliarobotics.com"
}

@article{manggala2024qa,
  title={Qa-calibration of language model confidence scores},
  author={Manggala, Putra and Mastakouri, Atalanti and Kirschbaum, Elke and Kasiviswanathan, Shiva Prasad and Ramdas, Aaditya},
  journal={arXiv preprint arXiv:2410.06615},
  year={2024}
}

@article{glenberg_grounding_2002,
	title = {Grounding language in action},
	volume = {9},
	issn = {1531-5320},
	url = {https://doi.org/10.3758/BF03196313},
	doi = {10.3758/BF03196313},
	language = {en},
	number = {3},
	urldate = {2025-09-12},
	journal = {Psychonomic Bulletin \& Review},
	author = {Glenberg, Arthur M. and Kaschak, Michael P.},
	month = sep,
	year = {2002},
	pages = {558--565},
}

@article{rizzolatti_language_1998,
	title = {Language within our grasp},
	volume = {21},
	issn = {0166-2236},
	doi = {10.1016/s0166-2236(98)01260-0},
	language = {eng},
	number = {5},
	journal = {Trends in Neurosciences},
	author = {Rizzolatti, G. and Arbib, M. A.},
	month = may,
	year = {1998},
	pmid = {9610880},
	pages = {188--194},
}

@article{pulvermuller_brain_2005,
	title = {Brain mechanisms linking language and action},
	volume = {6},
	copyright = {2005 Springer Nature Limited},
	issn = {1471-0048},
	url = {https://www.nature.com/articles/nrn1706},
	doi = {10.1038/nrn1706},
	language = {en},
	number = {7},
	urldate = {2025-09-12},
	journal = {Nature Reviews Neuroscience},
	author = {Pulvermüller, Friedemann},
	month = jul,
	year = {2005},
	note = {Publisher: Nature Publishing Group},
	pages = {576--582},
}
\bibliographystyle{IEEEtran}
}

\end{document}